\documentclass[letterpaper, 10 pt, conference]{ieeeconf}  

\IEEEoverridecommandlockouts                              

\usepackage{graphicx} 
\graphicspath{{figs/}}
\usepackage[caption=false,font=footnotesize]{subfig}
\usepackage{amsmath} 
\usepackage{amssymb}  
\usepackage{mathtools}
\mathtoolsset{showonlyrefs}
\let\labelindent\relax
\usepackage{enumitem}
\usepackage{url}
\usepackage{xcolor}
\usepackage{comment}
\usepackage{csquotes}

\DeclareMathOperator*{\argmin}{arg\!\min~}

\DeclareMathOperator*{\subjto}{subject\,to~}

\title{\LARGE \bf
What Can Robots Teach Us About The COVID-19 Pandemic? Interactive Demonstrations of Epidemiological Models\\Using a Swarm of Brushbots
}

\author{Gennaro Notomista and Siddharth Mayya
\thanks{
	\copyright 2022 IEEE.  Personal use of this material is permitted.  Permission from IEEE must be obtained for all other uses, in any current or future media, including reprinting/republishing this material for advertising or promotional purposes, creating new collective works, for resale or redistribution to servers or lists, or reuse of any copyrighted component of this work in other works.
}%
\thanks{This work was supported by the U.S. Embassy in Rome via the Alumni Small Grants Program 2020.}
\thanks{This work is not related to Amazon.}
\thanks{G. Notomista is with the Department of Electrical and Computer Engineering, University of Waterloo, Waterloo, ON, Canada {\tt\small gennaro.notomista@uwaterloo.ca}}%
\thanks{S. Mayya is with Amazon Robotics, North Reading, MA, USA {\tt\small mayya.siddharth@gmail.com}}%
}

\begin{document}

\maketitle
\thispagestyle{empty}
\pagestyle{empty}

\begin{abstract}

This paper describes the methodology and outcomes of a series of educational events conducted in 2021 which leveraged robot swarms to educate high-school and university students about epidemiological models and how they can inform societal and governmental policies. With a specific focus on the COVID-19 pandemic, the events consisted of 4 online and 3 in-person workshops where students had the chance to interact with a swarm of 20 custom-built brushbots---small-scale vibration-driven robots optimized for portability and robustness. Through the analysis of data collected during a post-event survey, this paper shows how the events positively impacted the students' views on the scientific method to guide real-world decision making, as well as their interest in robotics.
\end{abstract}

\section{Introduction} \label{sec:intro}
Robotic systems have been effectively employed in educational applications with the aim of increasing engagement and social interaction among youngsters, rehabilitation or therapy, as well as enhancing the overall learning experience~\cite{belpaeme2018social}. In particular, there are many examples in the existing literature where the use of robots has made the educational experience more engaging and enjoyable, thus supporting knowledge retention, and leading to an overall positive perception of the experience, e.g.,~\cite{wainer2007embodiment, powers2007comparing, li2015benefit}. \par 

Another technology has made a larger impact in shaping the modern classroom: the increasing popularity of social media and ubiquitous access to information has brought many benefits as well as risks. For instance, combating misinformation has been a topic of increased focus in the literature~\cite{adamic2016information,gradon2021countering,collins2021trends}, due to its far reaching impacts in society, as became clear during the COVID-19 pandemic. \par 

\begin{figure}[h!]
\centering
\includegraphics[width=0.95\linewidth]{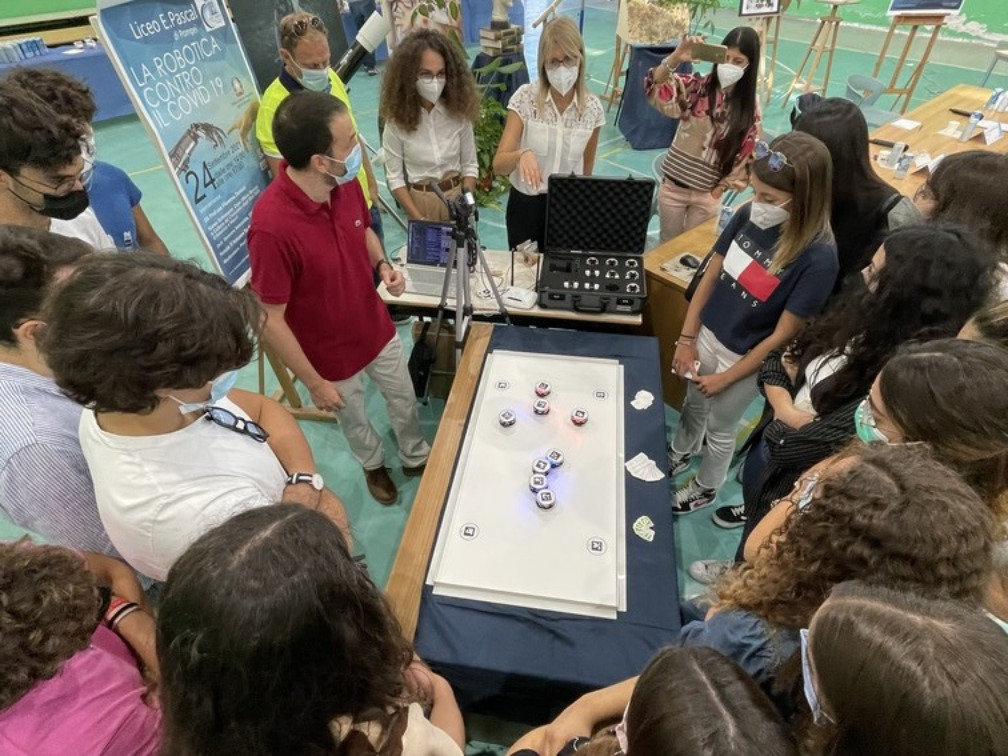}
\caption{High-school students at Liceo ``Ernesto Pascal'' in Pompei, Italy, looking at a group of 8 brushbots demonstrating a control policy to mitigate the spread of epidemics.}
\label{fig:pompeidemo}
\end{figure}
At the intersection of robotics-enabled education and increasing awareness of scientifically-backed societal and governmental decision-making, this paper describes and evaluates robotics educational events that we conducted in 2021, as part of the Alumni Small Grants Program\footnote{\url{https://it.usembassy.gov/it/alumni-small-grant-program-2020/}}, where we leveraged the educative and expressive power of swarm robotics systems to convey the importance of mathematical modeling in making real-world decisions and how they can inform social and governmental policies that can benefit society as a whole. We designed and conducted educational events in 2021, including 4 online and 3 in-person workshops for students as well as the general public, where participants had a chance to interact with a swarm of robots and observe experiments simulating the spread and control of communicable diseases like COVID-19. \par 

The focus for these events centered around two main questions:
\begin{enumerate}[label=(\roman*)] 
	\item What mathematical modeling tools exist that can allow us to predict the spread of diseases?
	\item How can we directly apply knowledge of these models to design policies to control their spread?
\end{enumerate}
As we demonstrate in the paper, the use of physically embodied robot swarms directly increases the efficacy of lessons on these topics. \par 

During the in-person workshops, we leveraged a swarm of \emph{brushbots}, which are custom-built vibration-driven robots that can operate on a portable platform (see Fig.~\ref{fig:pompeidemo}). Brushbots are suitable for educational applications thanks to their small size and robust construction~\cite{notomista2019study}. In particular, these robots can sustain collisions with each other~\cite{mayya2019non}, making them ideal for interactive engagement in the classroom. RGB LEDs mounted on the robots enable the swarm to display the simulation state to enhance legibility. For example, in the case of experiments on epidemiological models, different LED colors can be used to identify robots that are susceptible, infected, or recovered from a disease. Finally, a portable tabletop setup, along with compact wireless charging allows a swarm of 20 brushbots to be easily transported to any location for lessons and demonstrations. The impact that the robots had on the overall curriculum was evaluated by means of a survey conducted after the in-person workshops, whose results will be discussed and analyzed in this paper. \par 

The rest of this paper is organized as follows. Section~\ref{sec:lit} discusses related works dealing with robotics in education. Section~\ref{sec:swarm} outlines the design of the brushbot swarm and specific design decisions which make them amenable for use in educational applications. Section~\ref{sec:epidem} introduces a basic epidemiological model along with strategies to control  the spread of a disease. Section~\ref{sec:events} describes the 3 in-person events we organized at the end of 2021, and the various steps that were taken to enhance and optimize the learning experience. Section~\ref{sec:survey} outlines the results of the surveys we conducted after the events and Section~\ref{sec:disc} uses these results to discuss the efficacy of robot-centric learning in this topic. Section~\ref{sec:conc} concludes the paper. 

\section{Related Work} \label{sec:lit}
In this section, we briefly discuss relevant literature on the use of robotics systems in education. Robot systems are increasingly being used in education as tutors or peer learners, given their ability to increase cognitive engagement and, in some tasks, be as effective as human tutoring~\cite{belpaeme2018social}. Examples of social robots used in education are Keepon and Dragonbot, which are both animal-like, and human-like robots such as NAO, Wakamaru, and Robovie~\cite{konijn2020use}. In~\cite{kradolfer2014sociological}, the authors highlight the importance of good design practices in order to ensure the adoption of robots by teachers and enable creativity in students. Different modalities of interactions between users and the robots have been envisioned, including the use of haptic feedback via robots~\cite{kim2019swarmhaptics}, using robotic interfaces to drive up social engagement~\cite{saerbeck2010expressive}, and leveraging robot swarms as we do in this paper~\cite{vitanza2019robot}. We refer the reader to the following surveys~\cite{mubin2013review, karim2015review} for an overview of the various robots which have been used in educational settings and their trade-offs.

\section{Swarm of Brushbots}\label{sec:swarm}

In this section, we present the robots used throughout the organized workshops. In Section~\ref{subsec:brushbot}, we introduce the main features of the robots, while Section~\ref{subsec:swarm} illustrates the system setup used to control a swarm of brushbots to show the behavior of epidemiological models during the robot demonstrations.

\subsection{Individual Robot Design}
\label{subsec:brushbot}

\begin{figure}
\centering
\subfloat[][A brushbot with charging hat for wireless inductive charging.]{\label{subfig:brushbot}\includegraphics[width=0.49\linewidth]{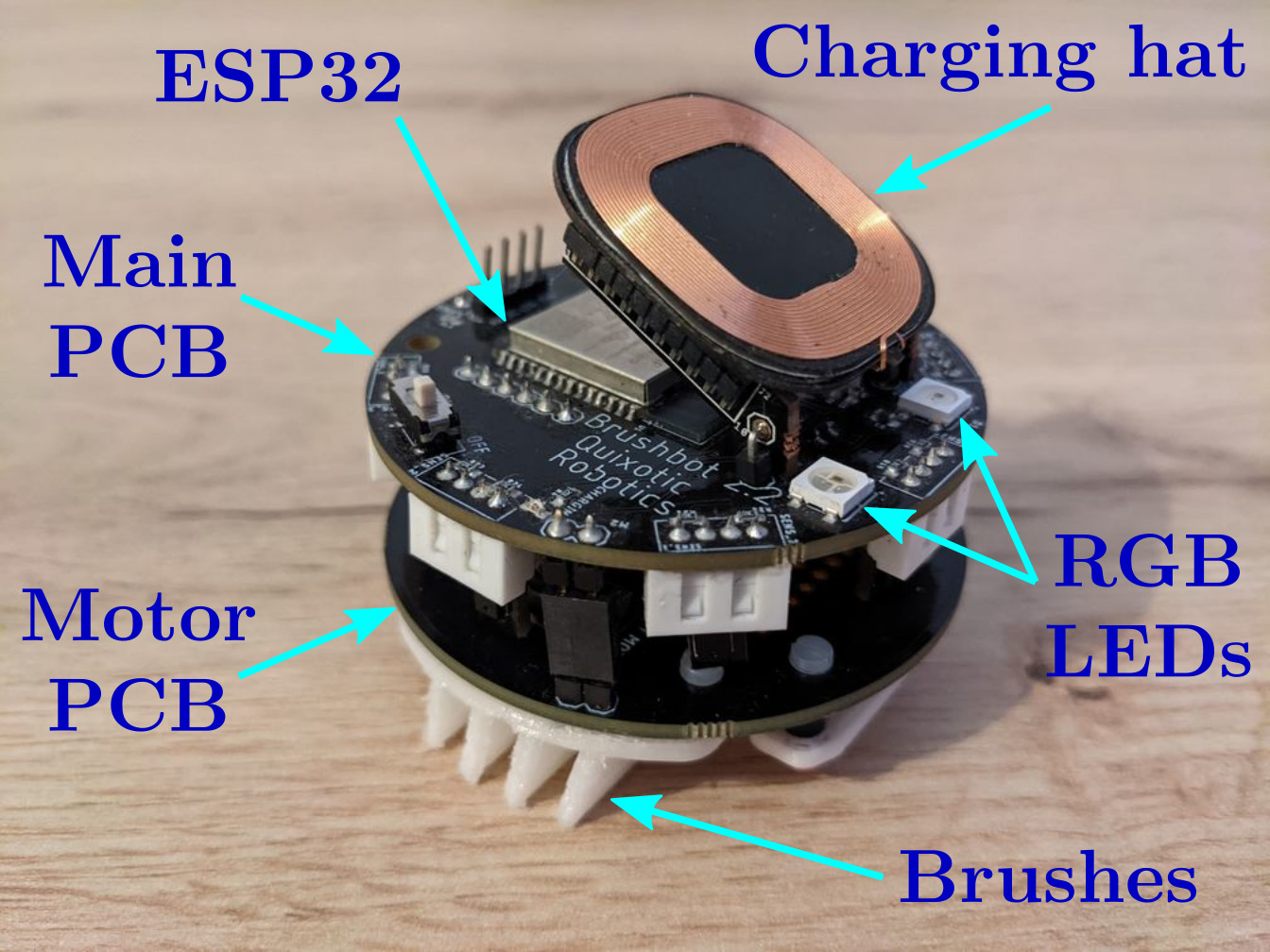}}\hspace{0.001cm}
\subfloat[][A group of 10 brushbots with tracking hats featuring fiducial markers. ]{\label{subfig:swarm}\includegraphics[width=0.49\linewidth]{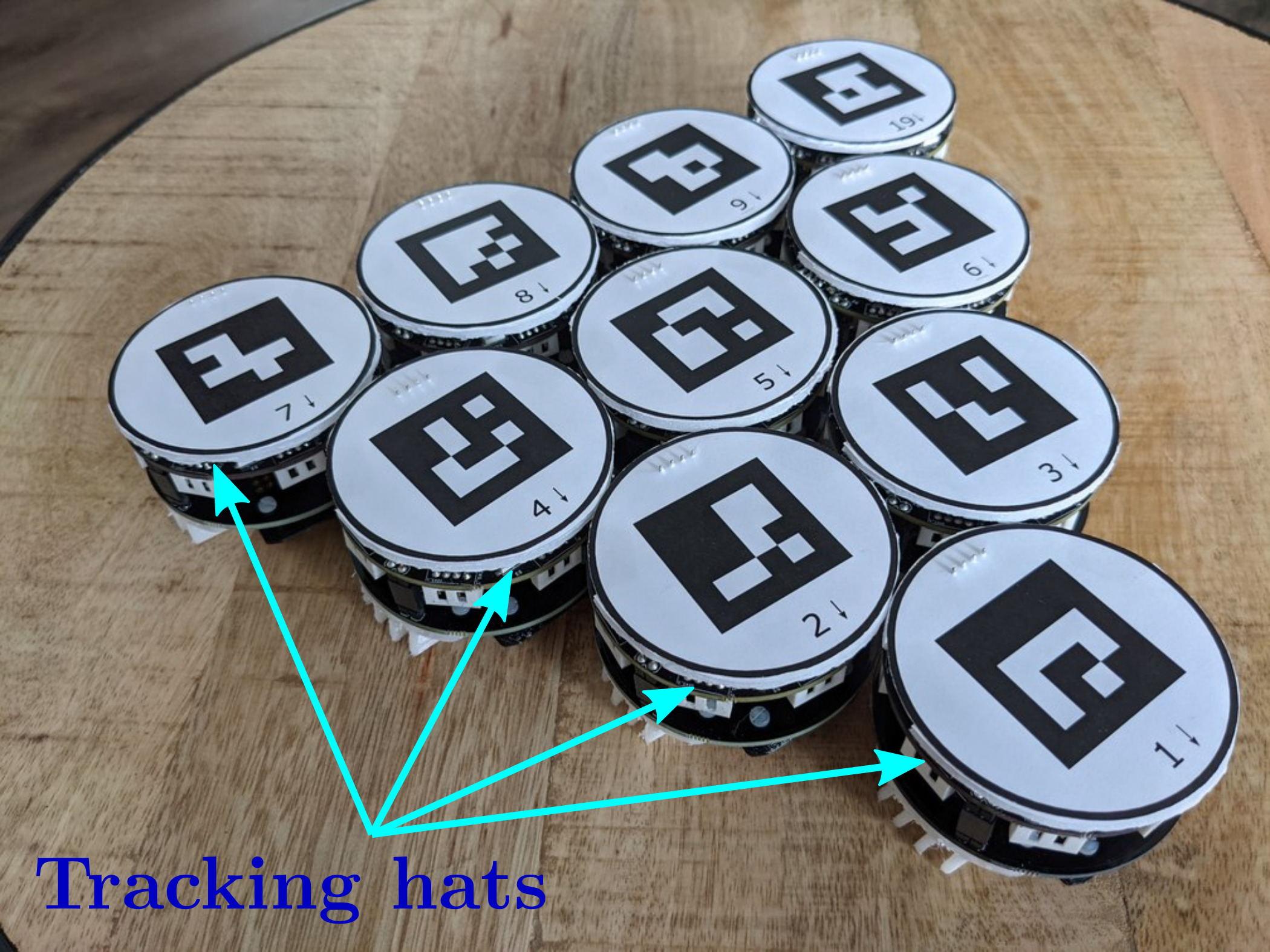}}
\caption{The brushbots used in the outreach events organized in the context of the project ``What can robots teach us about the COVID-19 pandemic''.}
\label{fig:brushbots}
\end{figure}
As discussed in the Introduction, brushbots are a class of vibration-driven robots that employ elastic elements---the brushes---to convert the energy of a vibration source into locomotion \cite{notomista2019study}. The specific instantiation of brushbots we used are shown in Fig.~\ref{fig:brushbots}. The robot is comprised of 2 custom circular printed circuit boards (the main PCB and the motor PCB) stacked on each other. The diameter of each board is 60mm. The brushes are 3D printed using a flexible material, the TPU 95A filament, and are mounted in a differential-drive configuration. On top of each brush, there is an eccentric rotating mass (ERM) motor which provides the required vibration. Compared to wheeled ground mobile robots, the brushbots are suitable for simulating a swarm of physically interacting particles, thanks to their locomotion principle, which allows them to resolve collisions by simply sliding along each other \cite{mayya2019non}. As will be explained in more detail in the next section, this allows us to employ them to demonstrate the spread of epidemics in a population of interacting individuals. More details on the modeling and control of the brushbots can be found in \cite{notomista2019study}.

The computational unit of the brushbots consists of an Espressif ESP32\footnote{\url{https://www.espressif.com/en/products/socs/esp32}}, programmed using Micropython\footnote{\url{https://micropython.org/}}. The firmware is automatically updated from a GitHub code repository\footnote{\url{https://github.com/brushbots/brushbot_firmware}} using Over-the-air (OTA) technologies. The brushbots are powered by a 3.7V 35C 220mAh LiPo battery, which provides them with an autonomy of about 45 minutes of average usage. Moreover, on-board power sensors allow the brushbots to monitor the power consumption and the state of the battery. Two programmable RGB LEDs are mounted on the main PCB at the front of the robot and can be used to visually convey information, an example of which will be given in Section~\ref{sec:events}. The brushbots feature a \textit{charging hat} (see Fig.~\ref{subfig:brushbot}), consisting of a coil connected on the main PCB, which can be placed near a wireless transmitter in order to activate the wireless recharging of the robot battery. A \textit{charging box} (visible in Fig.~\ref{fig:completesetup} in Subsection~\ref{subsec:swarm}) has been designed, which houses 10 wireless inductive chargers.

To localize the bruhbots on the plane where they locomote, dedicated \textit{tracking hats} have been designed. These hats consist of fiducial markers \cite{garrido2014automatic} printed on a circular white cardboard surface, which is mounted to the main PCB. Figure~\ref{subfig:swarm} shows 10 brushbots with 10 different fiducial markers to uniquely identify and track them.

\subsection{Robotic Swarm Design}
\label{subsec:swarm}

\begin{figure}
\centering
\includegraphics[width=0.95\linewidth]{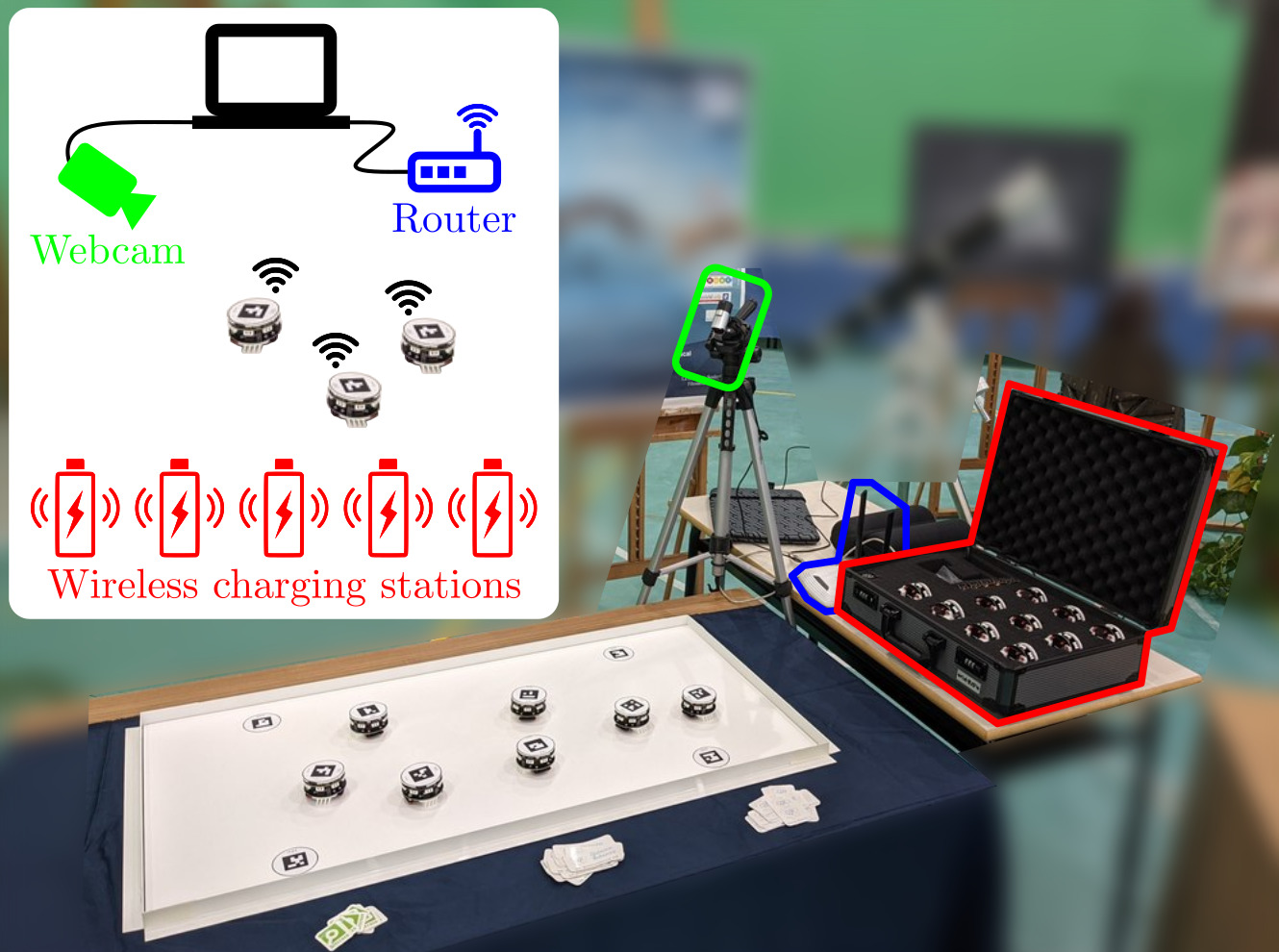}
\caption{Setup used for the brushbot demos during the outreach events. A PC running the swarm control software is connected to a webcam (green) via USB and to a router (blue) via Ethernet. The software receives sensor data from the brushbots and sends velocity input commands through WiFi using the UDP protocol. A charging box (red) housing 10 wireless inductive chargers is used to recharge the robots after each demo.}
\label{fig:completesetup}
\end{figure}

The swarm control architecture has been designed based on the Robot Operating System (ROS)\footnote{\url{https://www.ros.org/}} running on a dedicated PC. The main software component is the \textit{manager node}, that receives the poses of the brushbots from the \textit{tracker node}, which in turns computes them using a camera exploiting fiducial markers mounted on the robots. The manager node also receives sensor readings from the brushbots and sends velocity inputs to brushbots via UDP. For this, a dedicated router provides the wireless communication infrastructure: both the PC where the swarm control software runs and all the brushbots are on the same network with static IP addresses.

In order to control the robots to execute a desired task, an \textit{algorithm node} is implemented in the ROS network. This exchanges messages with the manager node by receiving robot poses and sending robot velocity inputs to be relayed to the robots. The hardware setup is illustrated in Fig.~\ref{fig:completesetup}.

The swarm of brushbots described in this section has been used during workshops open to high-school and university students, as well as the general public, to show the behavior and control of epidemics. In the next section, we briefly recall one of these models, the so-called SIR model, which has been implemented on the brushbots.

\section{Epidemiological Models Using Robot Swarms}\label{sec:epidem}

The dynamics of the spread of many viral diseases can be adequately captured by the so-called SIR model \cite{martcheva2015introduction}:
\begin{equation}
\label{eq:sir}
\begin{cases}
\dot S = -\beta I S\\
\dot I = \beta I S - \alpha I\\
\dot R = \alpha I,
\end{cases}
\end{equation}
where $S$, $I$, and $R$ denote the size of the populations of individuals who are susceptible, infected, and recovered, respectively. The total number of individuals in a population, $N$, is given by $N = S+I+R$. For short enough intervals of time, the spread of COVID-19 can be described accurately enough using the SIR dynamics \eqref{eq:sir}.

\begin{figure}
\centering
\includegraphics[width=\linewidth]{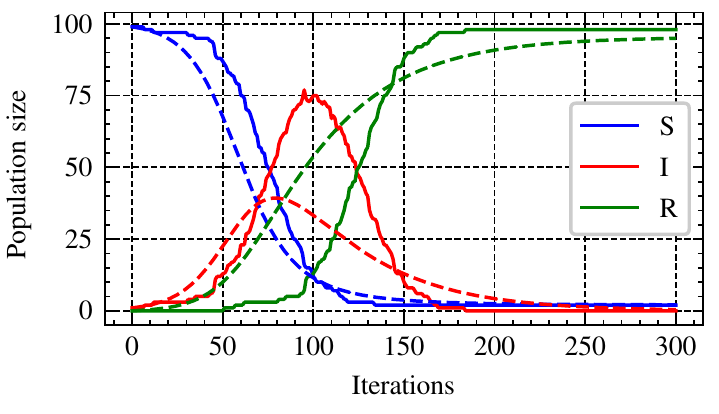}
\caption{Simulation results of the SIR model \eqref{eq:sir} obtained by numerically integrating Eq.~\ref{eq:sir} (with $\beta = 0.001$, $\alpha = 0.025$, and integration step equal to 1) and using a population of $100$ robots (with $P_\mathrm{infection} = 1$, $T_\mathrm{recover} = 50$, $d_\mathrm{thresh}=0.2$), respectively.}
\label{fig:sir}
\end{figure}
The parameter $\beta$ in \eqref{eq:sir} is evaluated as $\beta = P_\mathrm{infection} c$, where $P_\mathrm{infection}$ is the probability that a contact between an infected and a susceptible individual results in transmission, and $c$ is the per capita contact rate, directly proportional to the distance at which the disease can be transmitted. The parameter $\alpha$ is the recovery rate and it is inversely proportional to the time to recover, denoted by $T_\mathrm{recover}$. This correspondence between the parameters $\beta$ and $\alpha$ of the SIR model \eqref{eq:sir} and $P_\mathrm{infection}$ and $T_\mathrm{recover}$ allows us to simulate the SIR model using a swarm of brushbots randomly moving in a confined environment and exchanging the disease when they get in contact, i.e. when their distance falls below a threshold---the transmission distance $d_\mathrm{thresh}$. Figure~\ref{fig:sir} shows the comparison of simulation results of the SIR model obtained by solving \eqref{eq:sir} and by simulating $N=100$ robots randomly moving in a square environment of side length equal to $10$ units.

\subsection{Control of the Parameters of the SIR Model}
\label{subsec:control}

By varying the parameters of the model, certain objectives on the behavior of the epidemic can be obtained. For instance, during the outreach events organized in the context of this project, the participating students were given 3 control knobs ($P_\mathrm{infection}$, $T_\mathrm{recover}$ and $d_\mathrm{social}$) that they needed to tune in order to try to mitigate the spread.

$P_\mathrm{infection}$ and $T_\mathrm{recover}$ can be directly set as parameters of the simulation. Keeping the robots at a desired minimum social distance $d_\mathrm{social}$ from each other, on the other hand, is obtained by leveraging control barrier functions \cite{ames2019control}---a nonlinear control technique commonly employed to ensure the safety of dynamical systems. This entails letting the robots move with a velocity $u_1^\star, \ldots, u_N^\star$ given by the solution of the following optimization program:
\begin{equation}
\label{eq:qp}
\begin{aligned}
u_1^\star, \ldots, u_N^\star = \argmin_{u_1,\ldots,u_N} &\sum_{i\le N}\|u_i-\hat u_i\|^2\\
\subjto & \dot h(x_i,x_j,u_i,u_j) \ge -\alpha\left(h(x_i,x_j)\right)\\
& \hspace{6em}\forall i<j\le N,
\end{aligned}
\end{equation}
where $x_i$ denotes the position of robot $i$ on the plane, and $\hat u_i$ is the control input to drive robot $i$ randomly in a rectangular environment. The function $h$ is a control barrier function defined as follows:
\begin{equation}
h(x_i,x_j) = \|x_i-x_j\|^2-d_\mathrm{social}^2.
\end{equation}
The inequality constraint $\dot h(x_i,x_j,u_i,u_j) \ge -\alpha\left(h(x_i,x_j)\right)$---where $\alpha$ is any continuous, monotonically increasing function such that $\alpha(0)=0$---ensures that the velocity control input, $u_1^\star, \ldots, u_N^\star$, solutions of the optimization program \eqref{eq:qp} are able to keep the robots at least at a distance $d_\mathrm{social}$ from each other at any point in time.

The SIR model presented in this section has been implemented both in simulation and using a swarm of real brushbots during the workshops organized in the context of the project ``What can robots teach us about the COVID-19 pandemic'', the description of which is the focus of next section.

\section{Events and Methodology}\label{sec:events}

\begin{table*}
	\centering
	\caption{In-person events organized in the context of the project ``What can robots teach us about the COVID-19 pandemic''}
	\begin{tabular}{lllll}
		Venue & Location & Date & Attendees & Number of participants \\
		\hline
		High school Liceo ``Ernesto Pascal'' & Pompei & September 24, 2021 & High school students & 80 (approx.)\\
		Università degli Studi di Modena e Reggio Emilia & Reggio Emilia & November 23, 2021 & Graduate students & 40 (approx.)\\
		Public fair (\url{www.futuroremoto.eu}) & Napoli & November 27, 2021 & General public & 80 (approx.)
	\end{tabular}
	\label{tab:events}
\end{table*}

In the context of the project ``What can robots teach us about the COVID-19 pandemic'', two types of events have been organized:
\begin{enumerate}[label=(\roman*)]
\item A series of 4 online workshops in April 2021
\item A series of 3 in-person workshops with robot demonstrations in September and November 2021, where the brushbots have been employed to show the evolution and control of epidemics.
\end{enumerate}
More pictures and resources about the events are available online at \url{https://www.quixoticrobotics.org/#events}.

\subsection{Online Events}

The purpose of the online events was to prepare the students for the in-person workshops, by giving them an overview of the mathematical epidemiological models, including the SIR model described in Section~\ref{sec:epidem}.

\begin{figure}
\centering
\includegraphics[width=0.8\linewidth,trim={0px 290px 0px 0px},clip]{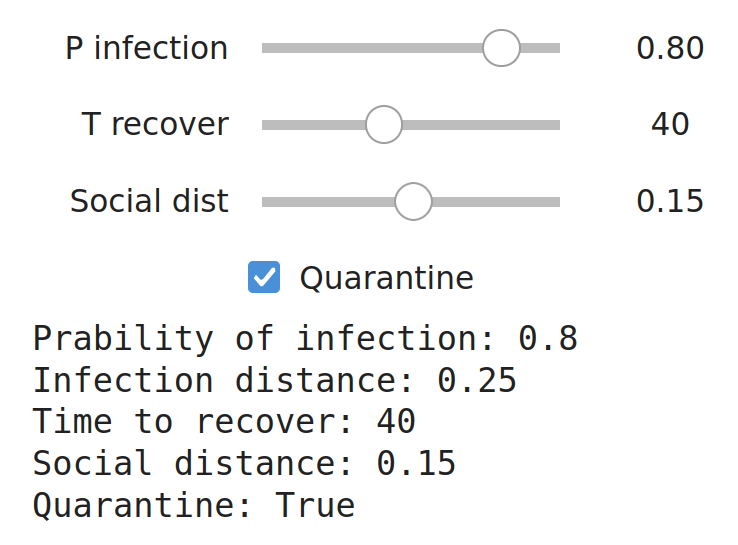}
\caption{Screenshot of a section of the Google Colab used to experiment with controlling the SIR epidemiological model during a coding lab session. $P_\mathrm{infection}$, $T_\mathrm{recover}$, and $d_\mathrm{social}$ can be chosen using sliders before running the simulation. The links to the Google Colabs used during the online events are available at \protect\url{https://www.quixoticrobotics.org/robots_and_covid.html\#la_robotica_contro_il_covid}.}
\label{fig:controlknobs}
\end{figure}
One of the online workshops has been held in the form of a coding lab, where the Robotarium simulator \cite{wilson2020robotarium} has been presented to simulate the control of epidemics.
The three parameters $P_\mathrm{infection}$, $T_\mathrm{recover}$, $d_\mathrm{social}$ used to simulate the SIR model on a swarm of robots were given as control knobs to the students in a game programmed using Python and Google Colab (see Fig.~\ref{fig:controlknobs}). The goal of the game is to achieve the highest score $s$ defined as follows:
\begin{equation}
s = 100\,(s_h + s_f + s_p + s_e),
\end{equation}
where
\begin{equation}
\begin{aligned}	
s_h &= \frac{10}{1+\max{\{I(t)\}_{t\le T_\mathrm{max}}}} + \frac{10}{1+T_\mathrm{recover}}\\[1em]
s_f &= P_\mathrm{infection} +  \frac{T_{max}}{\sum_{t=0}^{T_\mathrm{max}}\|u^\star-\hat u\|}\\[1em]
s_p &= \frac{10}{1+\max{\{I(t)\}_{t\le T_\mathrm{max}}}}\\[1em]
s_e &= 0.1\,T_\mathrm{recover} + 10\,(1-P_\mathrm{infection})\\
&\quad + \frac{10}{1+\max{\{I(t)\}_{t\le T_\mathrm{max}}}}.
\end{aligned}
\end{equation}

The online workshops have been attended by about 100 students on average, with a peak of 118, most of whom also took part in the in-person events described in the following section. Moreover, several students have designed their final high school project around the topics of the workshops. One of these students is Nunzia D'Antuono, alumna of the high school Liceo ``Ernesto Pascal'', who commented about her whole experience as follows:
\begin{displayquote}
        \itshape
        [...] What sparked my interest so significantly, in addition to graph theory, the processes of spread of the virus and the ways of using robotics in the health and prevention field, was the virtual simulation of the infection [...] The project was so exciting to lead me to use it for the high school final exam project. In fact, using robots in the Robotarium, the virtual simulation turned into real, achieving successfully the expected outcome. The experience was really constructive and stimulating, absolutely to propose again.
\end{displayquote}

\subsection{In-person Events}

Three in-person workshops featuring swarms of brushbots to demonstrate the control of epidemics have been organized. Details regarding locations and dates of the events are reported in Table~\ref{tab:events}. The participants to the workshops came from schools (5th-12th grade), universities (undergraduate and graduate), as well as the general public. During the events, we distributed USB flash drives containing the open-source brushbot simulator, designed to minimize the gap between simulation and real experiments with the brushbots. This allowed us to keep working remotely with some of the students who attended the in-person workshops and who are interested in further developing epidemiological control algorithms using real brushbots.

\begin{figure}
\centering
\includegraphics[width=0.95\linewidth,trim={0px 0px 0px 250px},clip]{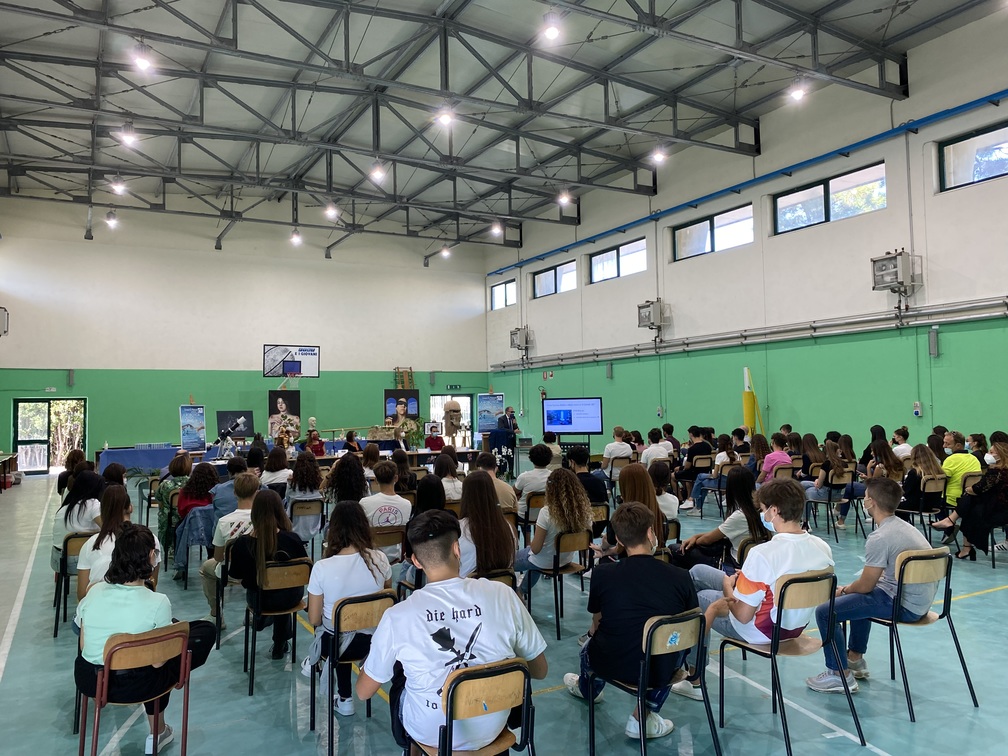}
\caption{In-person event at the high school Liceo ``Ernesto Pascal''}
\label{fig:pompeicrowd}
\end{figure}
The workshops organized in Pompei---where we conducted the survey presented in Sections~\ref{sec:survey} and \ref{sec:disc}---was held at the high school Liceo ``Ernesto Pascal'' (Fig.~\ref{fig:pompeicrowd}). Approximately 80 12th grade students participated to the event. The event was also attended by the Mayor of Pompei and by Karen Schinnerer, Public Affairs Officer at the U.S. Consulate General in Napoli, who commented on the series of workshops as follows:
\begin{displayquote}
        \itshape
        [...] Sharing this key collaborative research with future generations of high school and university students demonstrates the utility of international exchanges and how concrete research connects to today’s global economy. We know these outreach events at schools, universities and local science fairs reach new audiences often leading students to explore new fields of study [...]
\end{displayquote}

\begin{figure}
\centering
\includegraphics[width=0.95\linewidth]{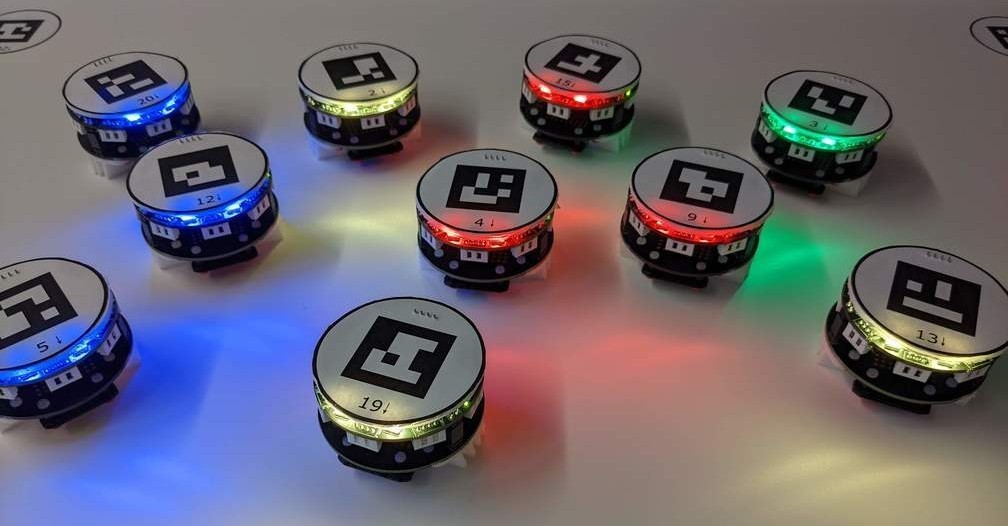}
\caption{RGB LEDs of the brushbots used to identify individuals belong to the S (susceptible, blue), I (infected, red), R (recovered, green) classes of the SIR model. The color yellow is used to identify vaccinated individuals.}
\label{fig:leds}
\end{figure}
The setup of the in-person events in Pompei is the one shown in Fig.~\ref{fig:completesetup}. The brushbots move on a rectangular flat and smooth portable surface following random trajectories, simulating the spread of an epidemic using the SIR model described in Section~\ref{sec:epidem}. Their state---i.e. whether they are susceptible (S), infected (I), or recovered (R)---is displayed using different colors of the built-in RGB LEDs, as shown in Fig.~\ref{fig:leds}. The students were able to see the effects of different control policies obtained by varying the parameters $P_\mathrm{infection}$, $T_\mathrm{recover}$ and $d_\mathrm{social}$, as described in Section~\ref{subsec:control}.

The impact of the in-person workshops has been evaluated by means of a survey we conducted after the event. In the next section, we introduce the post-event survey, whose results are discussed in detail in Section~\ref{sec:disc}.

\section{Post-event Survey}\label{sec:survey}

\begin{figure*}
\centering
\subfloat[][Importance of the events on students' learning about epidemiological models (Q1)]{\includegraphics[width=0.35\linewidth]{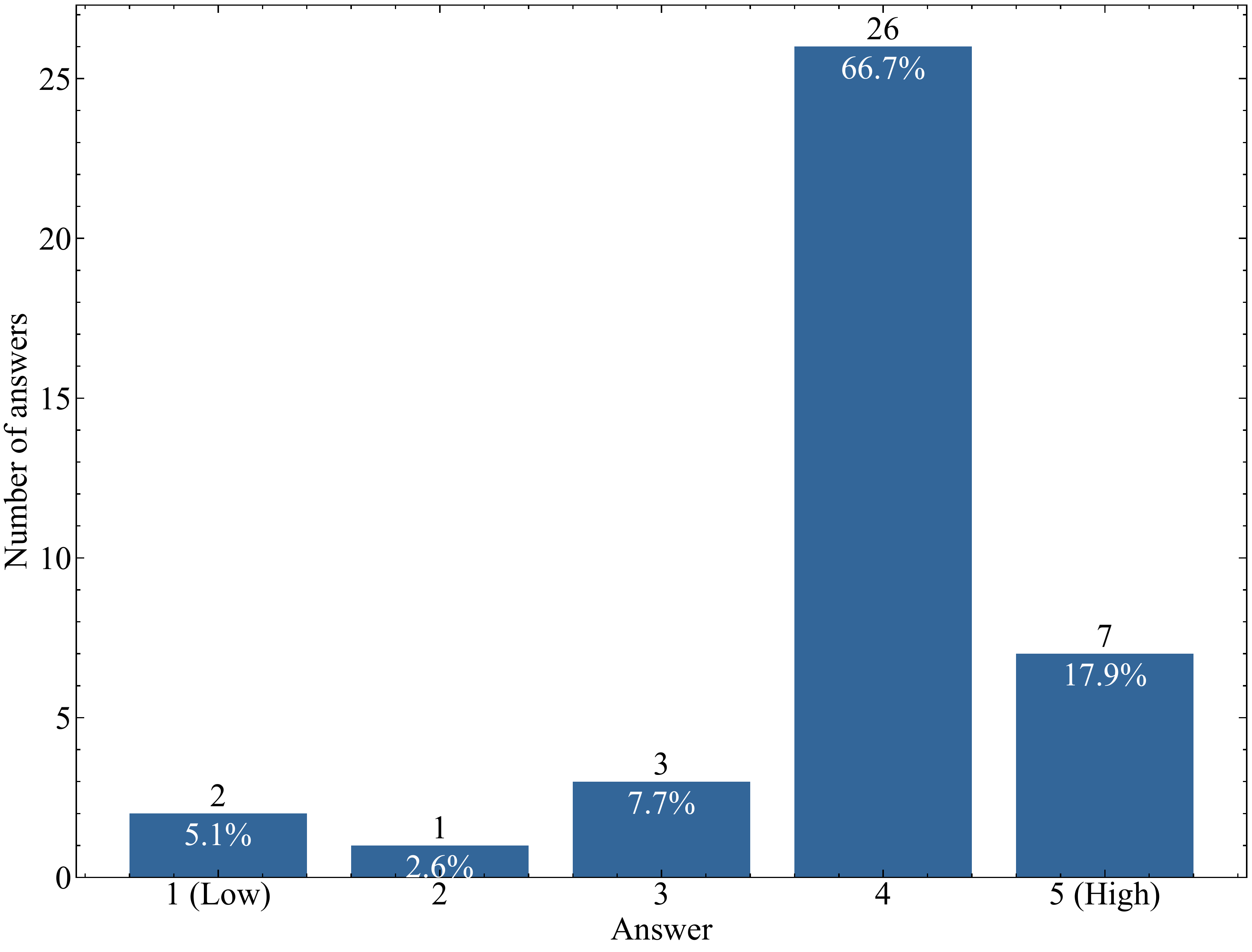}}\hfill
\subfloat[][Impact of the event on students' attitude towards science (Q2)]{\includegraphics[width=0.35\linewidth]{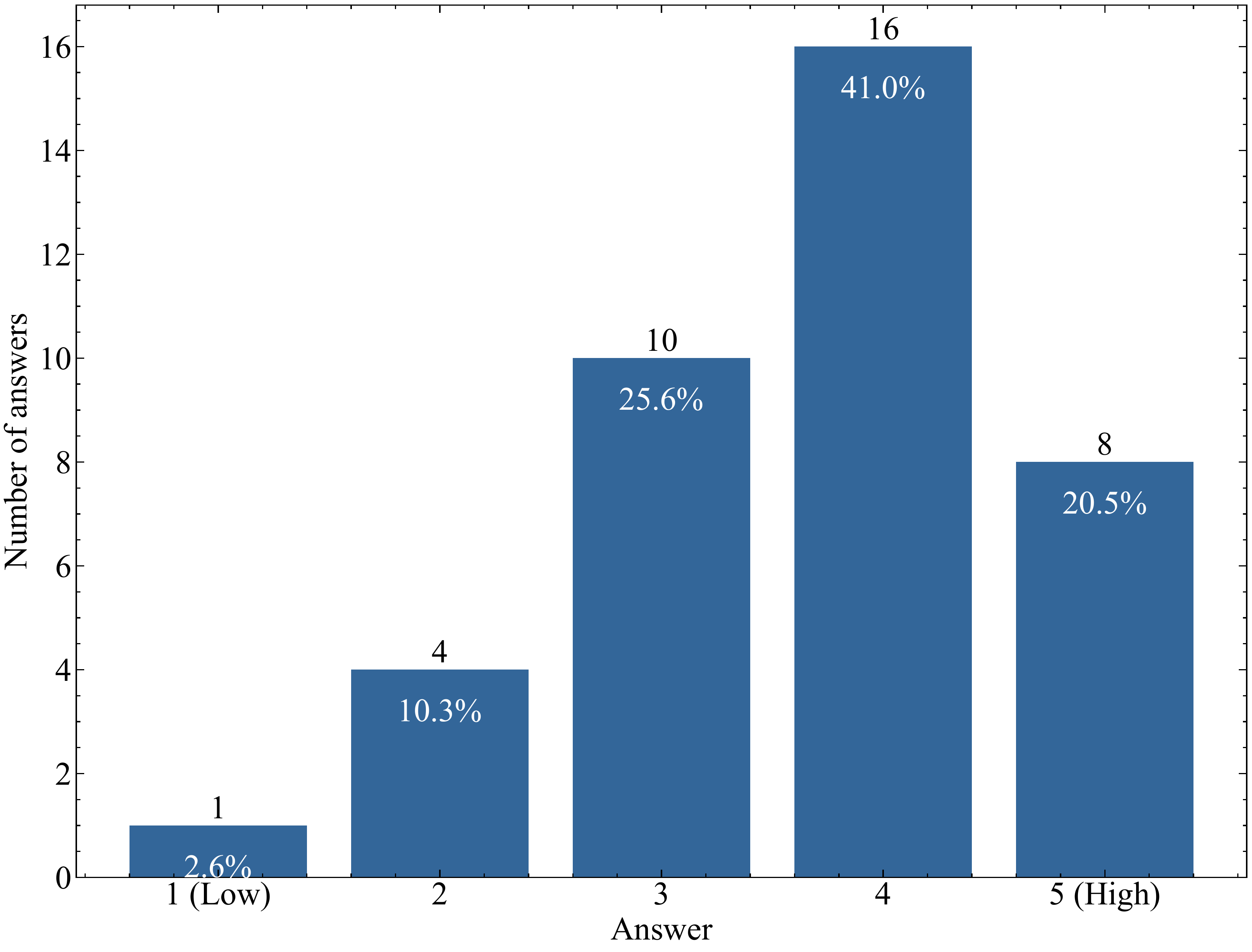}}\hfill
\subfloat[][Do students think science can help solve topical social issues? (Q3)]{\includegraphics[width=0.28\linewidth]{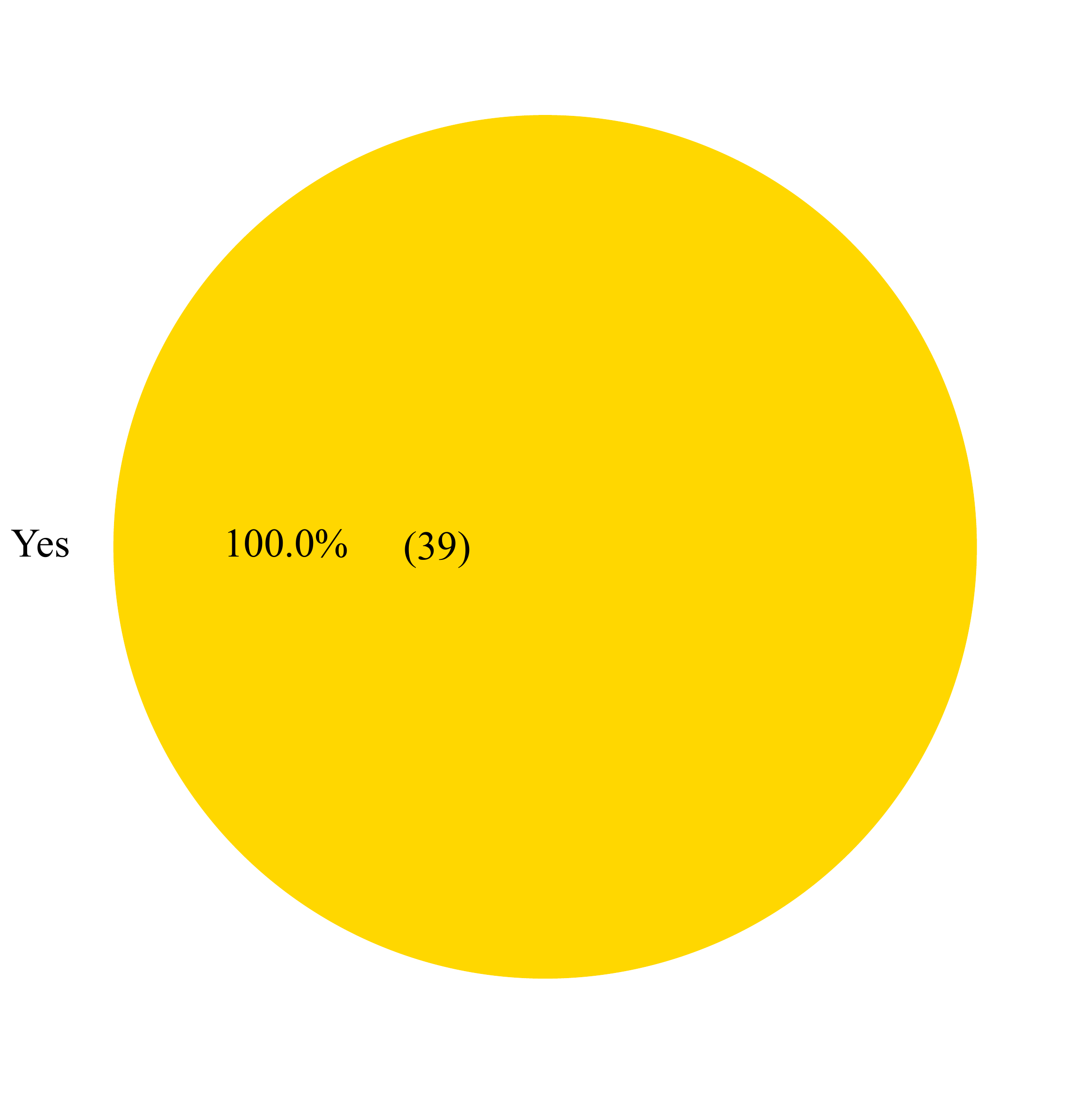}}\\
\subfloat[][Didactic usefulness of robots in understanding epidemiological models (Q4)]{\includegraphics[width=0.35\linewidth]{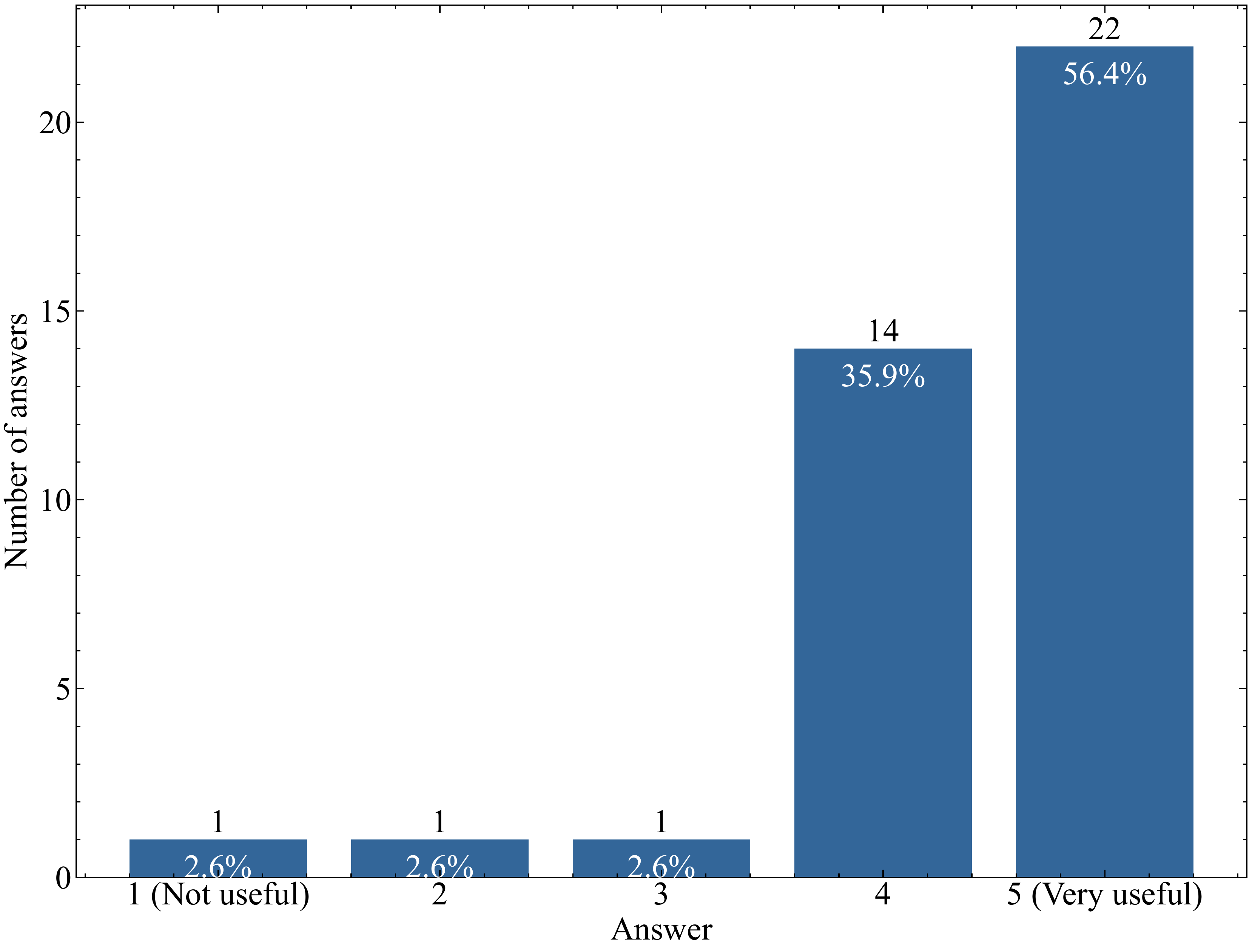}}\hfill
\subfloat[][Didactic usefulness of robots in understanding the control of epidemics (Q5)]{\includegraphics[width=0.36\linewidth]{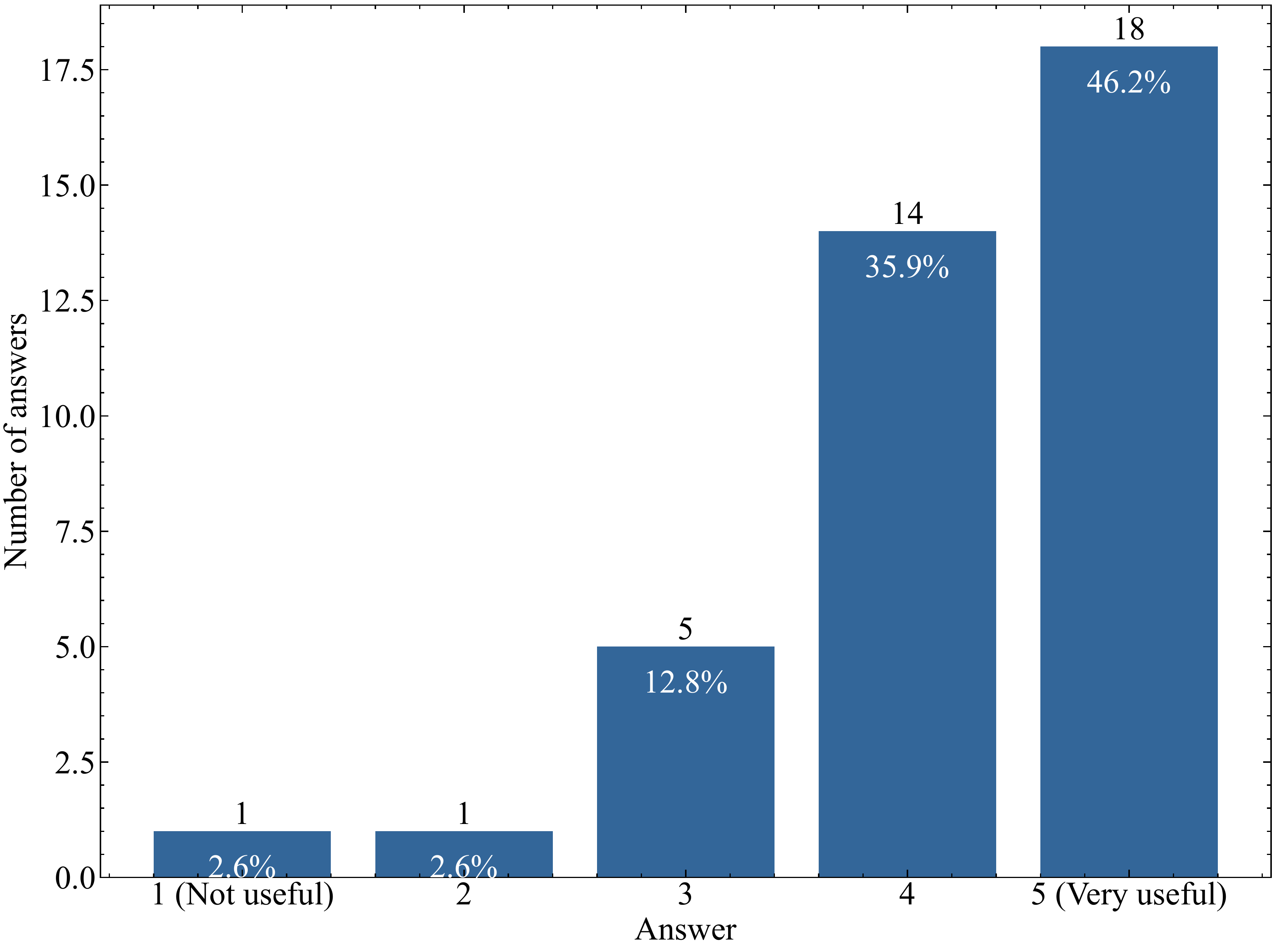}}\hfill
\subfloat[][Are students willing to organize similar events in the future? (Q6)]{\includegraphics[width=0.28\linewidth]{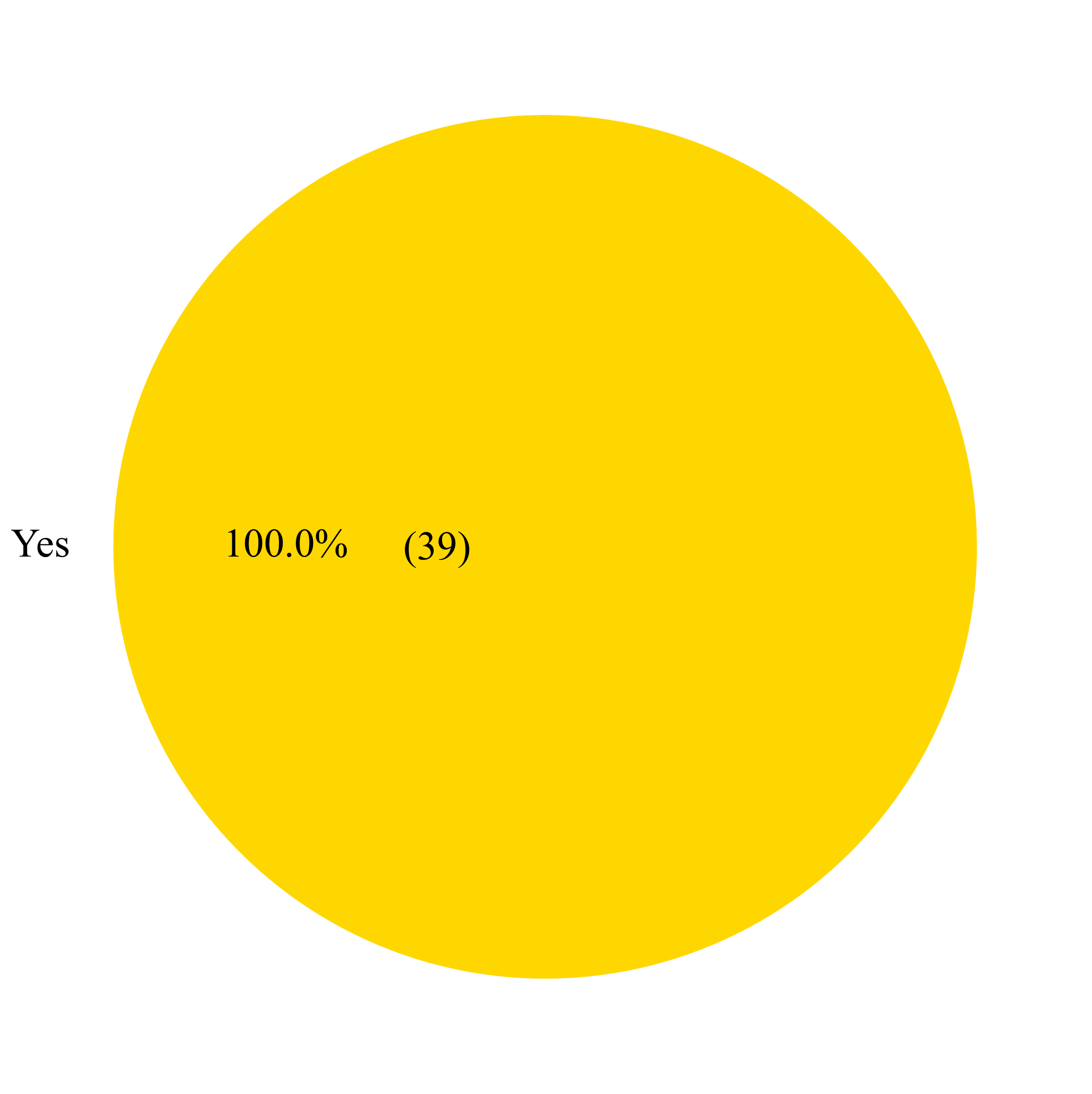}}\\
\subfloat[][Effort students would put in organizing similar events in the future (Q7)]{\includegraphics[width=0.35\linewidth]{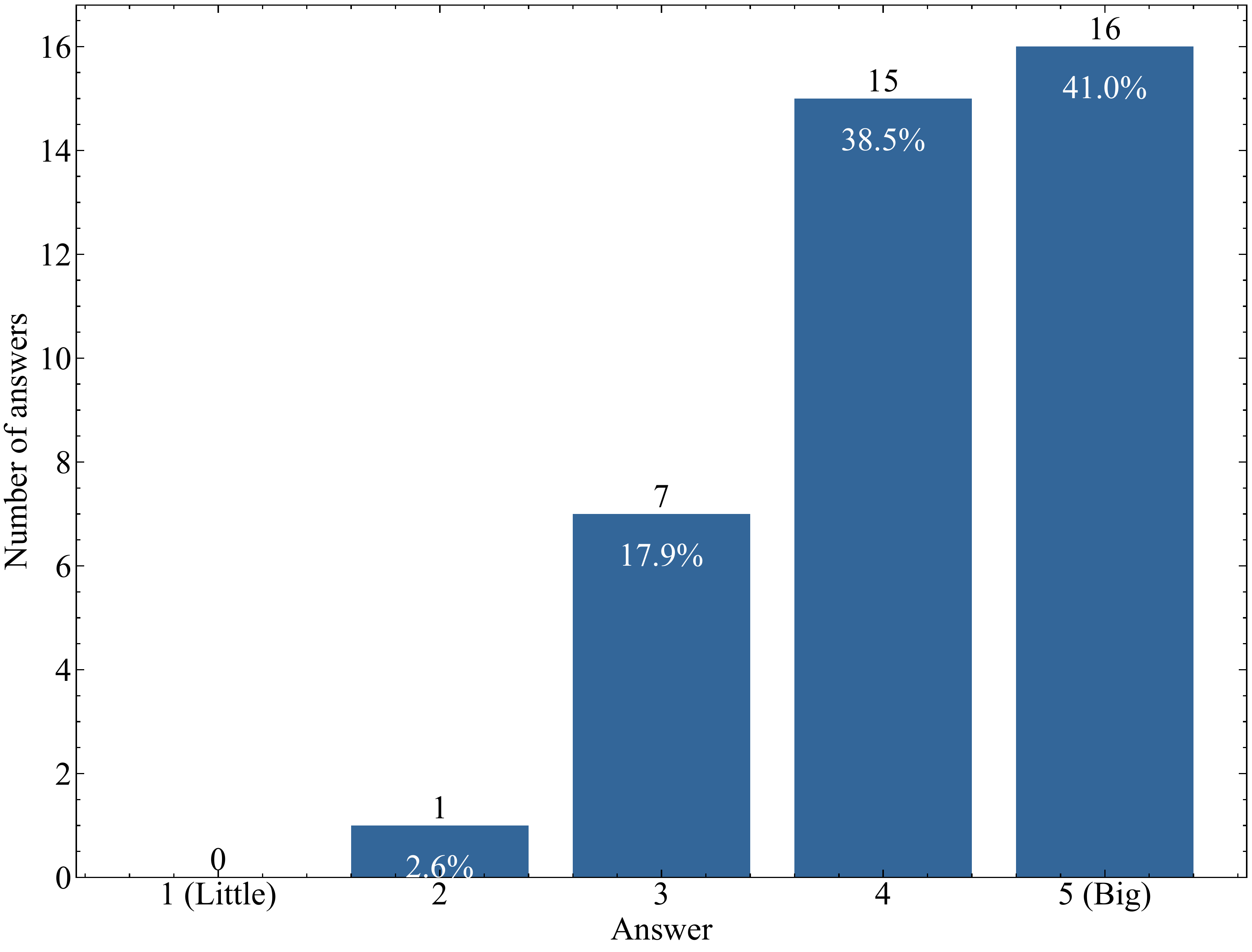}}\qquad
\subfloat[][Change in students' enthusiasm towards robotics before and after the event (Q8)]{\includegraphics[width=0.3\linewidth]{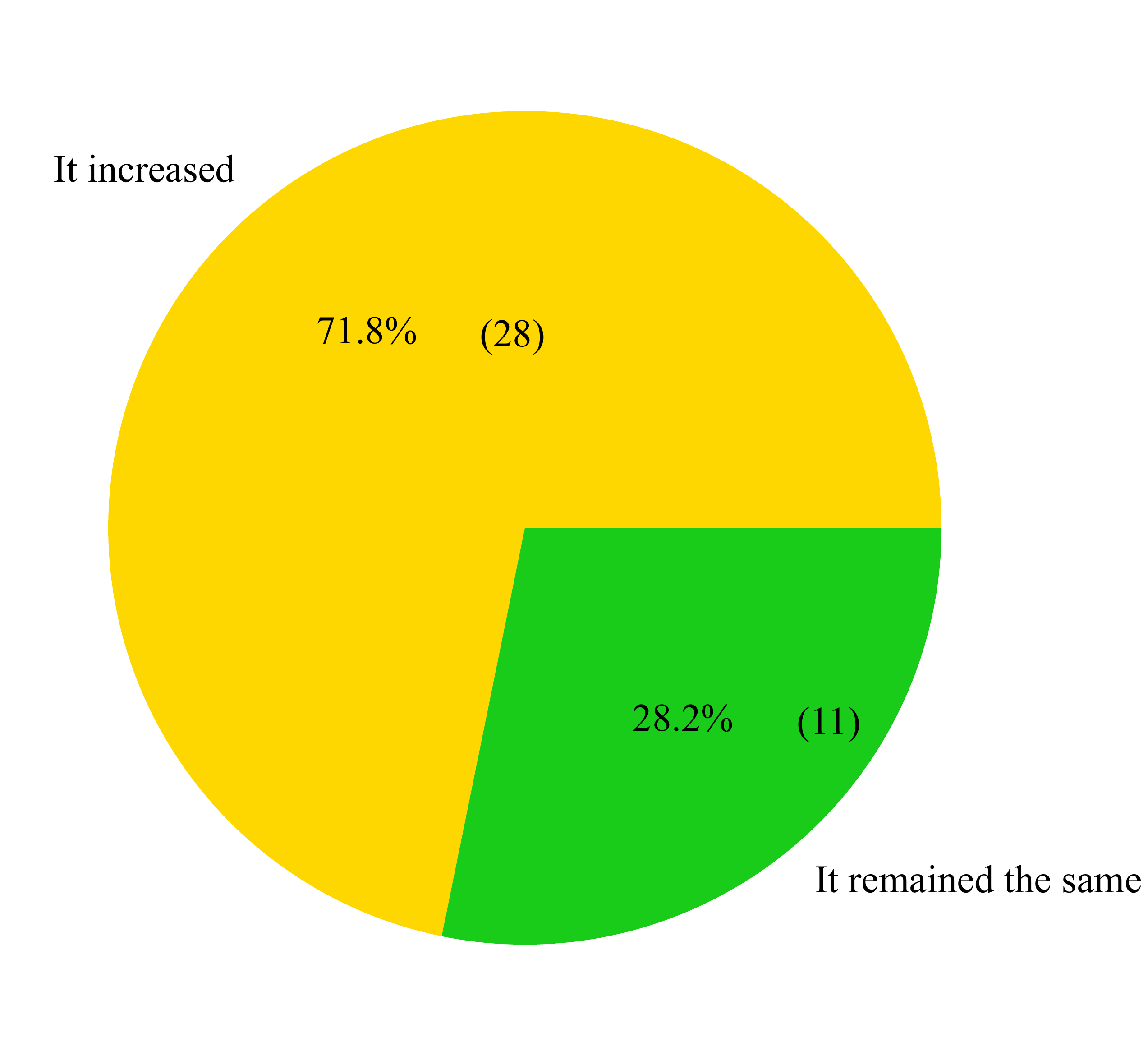}}
\caption{Responses of the survey}
\label{fig:Q}
\end{figure*}

A survey has been conducted with the following objective: \textit{Evaluating the effectiveness of using robots in increasing awareness of people with respect to topical social issues}. In particular, we were interested in quantifying how much impact the in-person workshops featuring swarms of brushbots had in positively changing people's attitude towards science and the scientific method, as well as the use of robotics to tackle social problems. We asked the students of the high school Liceo ``Ernesto Pascal'' in Pompei to answer the following 8 questions:
\begin{enumerate}[label=(Q\arabic*)]
\item How much would you rate the importance that the events of ``What can robots teach us about the COVID-19 pandemic'' had in making you learn about the use of mathematics to model, predict, and control epidemic diffusion? (1: Low to 5: High)
\item How much would you rate the impact of the events of ``What can robots teach us about the COVID-19 pandemic'' on your attitude towards science and the scientific method applied to topical social issues? (1: Low to 5: High)
\item Do you think science and the scientific method can help us solve topical social issues? (Yes or No)
\item How much would you rate the usefulness of employing real robots to illustrate the behavior of epidemiological models, like the SIR? (1: Not useful to 5: Very useful)
\item How much would you rate the usefulness of employing real robots to illustrate the effectiveness of epidemiological control policies, such social distancing, masking, and vaccines? (1: Not useful to 5: Very useful)
\item Would you enjoy organizing events like those of ``What can robots teach us about the COVID-19 pandemic'' aimed to promote the use of science and technology to increase the awareness on topical social issues? (Yes or No)
\item How much effort would you be willing to put into the organization of events like those of ``What can robots teach us about the COVID-19 pandemic''? (1: Little to 5: Big)
\item After participating in the events of ``What can robots teach us about the COVID-19 pandemic'', how did your interest about robotics in general change? (It increased, It remained the same, or It decreased)
\end{enumerate}
We collected 39 anonymous answers from students whose age ranges from 18 to 19 years old. The statistics of the responses to the survey are reported in Fig.~\ref{fig:Q} and will be discussed in the next section.

\section{Discussion}\label{sec:disc}

Question Q1 is concerned with the knowledge of epidemiological models. The responses to this question show that, despite the significant amount of news about COVID-19 which have been flowing to us via different media channels for the past 2 years, most students (84.6\%) rated with at least 4 out of 5 the importance of the organized events for learning about the use of mathematical models to predict the behavior of epidemics. Given these results for question Q1, the responses to Q2 and Q3 confirm that the events had a positive impact in changing students' attitude towards science and the scientific method applied to topical social issues, such as the control of epidemics.

An important aspect which we were interested in verifying is the role that robotics can play in outreach events in strengthening the conveyance of a desired concept. The responses to Q4 and Q5 show that 92.3\% and 81.1\% of the students think that employing real robots is useful to illustrate the behavior and the control, respectively, of epidemiological models. This demonstrates the potential effectiveness of the use of robotics to convey a desired message.

Besides the effectiveness of robots and robotics in conveying a message, we were interested in understanding whether such outreach events featuring robot demos inspire students to themselves participate in such activities (hence creating a cascading positive effect). The responses to Q6 and Q7 show that almost the majority of the students (97.4\%) would be inclined to put effort in the organization of events like those described in this work.

Finally, the responses to question Q8 show that in 71.8\% of the students the interest in robotics has increased after attending the organized events. This, besides the positive impact in conveying the desired message, shows that events featuring robots have the potential of increasing students' interest in robotics as a field.

Overall, the results of the survey were positive enough to motivate the organization of similar events at larger scales with the objectives of counteracting misinformation and further evaluating the ramifications of robotics in education.

\section{Conclusions}\label{sec:conc}
This paper described the preparation and organization of educational events at the intersection of epidemiology, scientific modeling, and robotics. The target audience was high-school and university students who would benefit from the increased cognitive engagement that mobile robots bring to the classroom. A portable and modular robot swarm design improved the effectiveness of the events, as confirmed by a post-event survey which demonstrated that more interactive events are warranted to engage the public in science-backed societal and governmental decision-making processes.






\section*{ACKNOWLEDGMENT}

The authors would like to acknowledge the help of Mrs. Emilia Letterese, English teacher of the high school Liceo ``Ernesto Pascal'', for translating the questions of the survey from English to Italian, as well as the comments of the student Nunzia D'Antuono from Italian to English.


\bibliographystyle{IEEEtran}
\bibliography{bib/references.bib}

\begin{thebibliography}{10}
\providecommand{\url}[1]{#1}
\csname url@samestyle\endcsname
\providecommand{\newblock}{\relax}
\providecommand{\bibinfo}[2]{#2}
\providecommand{\BIBentrySTDinterwordspacing}{\spaceskip=0pt\relax}
\providecommand{\BIBentryALTinterwordstretchfactor}{4}
\providecommand{\BIBentryALTinterwordspacing}{\spaceskip=\fontdimen2\font plus
\BIBentryALTinterwordstretchfactor\fontdimen3\font minus
  \fontdimen4\font\relax}
\providecommand{\BIBforeignlanguage}[2]{{%
\expandafter\ifx\csname l@#1\endcsname\relax
\typeout{** WARNING: IEEEtran.bst: No hyphenation pattern has been}%
\typeout{** loaded for the language `#1'. Using the pattern for}%
\typeout{** the default language instead.}%
\else
\language=\csname l@#1\endcsname
\fi
#2}}
\providecommand{\BIBdecl}{\relax}
\BIBdecl

\bibitem{belpaeme2018social}
T.~Belpaeme, J.~Kennedy, A.~Ramachandran, B.~Scassellati, and F.~Tanaka,
  ``Social robots for education: A review,'' \emph{Science robotics}, vol.~3,
  no.~21, p. eaat5954, 2018.

\bibitem{wainer2007embodiment}
J.~Wainer, D.~J. Feil-Seifer, D.~A. Shell, and M.~J. Mataric, ``Embodiment and
  human-robot interaction: A task-based perspective,'' in \emph{RO-MAN 2007-The
  16th IEEE International Symposium on Robot and Human Interactive
  Communication}.\hskip 1em plus 0.5em minus 0.4em\relax IEEE, 2007, pp.
  872--877.

\bibitem{powers2007comparing}
A.~Powers, S.~Kiesler, S.~Fussell, and C.~Torrey, ``Comparing a computer agent
  with a humanoid robot,'' in \emph{Proceedings of the ACM/IEEE international
  conference on Human-robot interaction}, 2007, pp. 145--152.

\bibitem{li2015benefit}
J.~Li, ``The benefit of being physically present: A survey of experimental
  works comparing copresent robots, telepresent robots and virtual agents,''
  \emph{International Journal of Human-Computer Studies}, vol.~77, pp. 23--37,
  2015.

\bibitem{adamic2016information}
L.~A. Adamic, T.~M. Lento, E.~Adar, and P.~C. Ng, ``Information evolution in
  social networks,'' in \emph{Proceedings of the ninth ACM international
  conference on web search and data mining}, 2016, pp. 473--482.

\bibitem{gradon2021countering}
K.~T. Grado{\'n}, J.~A. Ho{\l}yst, W.~R. Moy, J.~Sienkiewicz, and K.~Suchecki,
  ``Countering misinformation: A multidisciplinary approach,'' \emph{Big Data
  \& Society}, vol.~8, no.~1, 2021.

\bibitem{collins2021trends}
B.~Collins, D.~T. Hoang, N.~T. Nguyen, and D.~Hwang, ``Trends in combating fake
  news on social media--a survey,'' \emph{Journal of Information and
  Telecommunication}, vol.~5, no.~2, pp. 247--266, 2021.

\bibitem{notomista2019study}
G.~Notomista, S.~Mayya, A.~Mazumdar, S.~Hutchinson, and M.~Egerstedt, ``A study
  of a class of vibration-driven robots: Modeling, analysis, control and design
  of the brushbot,'' in \emph{2019 IEEE/RSJ International Conference on
  Intelligent Robots and Systems (IROS)}.\hskip 1em plus 0.5em minus
  0.4em\relax IEEE, 2019, pp. 5101--5106.

\bibitem{mayya2019non}
S.~Mayya, G.~Notomista, D.~Shell, S.~Hutchinson, and M.~Egerstedt,
  ``Non-uniform robot densities in vibration driven swarms using phase
  separation theory,'' in \emph{2019 IEEE/RSJ International Conference on
  Intelligent Robots and Systems (IROS)}.\hskip 1em plus 0.5em minus
  0.4em\relax IEEE, 2019, pp. 4106--4112.

\bibitem{konijn2020use}
E.~A. Konijn, M.~Smakman, and R.~van~den Berghe, ``Use of robots in
  education,'' \emph{The International Encyclopedia of Media Psychology}, pp.
  1--8, 2020.

\bibitem{kradolfer2014sociological}
S.~Kradolfer, S.~Dubois, F.~Riedo, F.~Mondada, and F.~Fassa, ``A sociological
  contribution to understanding the use of robots in schools: the thymio
  robot,'' in \emph{International Conference on Social Robotics}.\hskip 1em
  plus 0.5em minus 0.4em\relax Springer, 2014, pp. 217--228.

\bibitem{kim2019swarmhaptics}
L.~H. Kim and S.~Follmer, ``Swarmhaptics: Haptic display with swarm robots,''
  in \emph{Proceedings of the 2019 CHI conference on human factors in computing
  systems}, 2019, pp. 1--13.

\bibitem{saerbeck2010expressive}
M.~Saerbeck, T.~Schut, C.~Bartneck, and M.~D. Janse, ``Expressive robots in
  education: varying the degree of social supportive behavior of a robotic
  tutor,'' in \emph{Proceedings of the SIGCHI conference on human factors in
  computing systems}, 2010, pp. 1613--1622.

\bibitem{vitanza2019robot}
A.~Vitanza, P.~Rossetti, F.~Mondada, and V.~Trianni, ``Robot swarms as an
  educational tool: The thymio’s way,'' \emph{International Journal of
  Advanced Robotic Systems}, vol.~16, no.~1, 2019.

\bibitem{mubin2013review}
O.~Mubin, C.~J. Stevens, S.~Shahid, A.~Al~Mahmud, and J.-J. Dong, ``A review of
  the applicability of robots in education,'' \emph{Journal of Technology in
  Education and Learning}, vol.~1, no. 209-0015, p.~13, 2013.

\bibitem{karim2015review}
M.~E. Karim, S.~Lemaignan, and F.~Mondada, ``A review: Can robots reshape k-12
  stem education?'' in \emph{2015 IEEE international workshop on Advanced
  robotics and its social impacts (ARSO)}.\hskip 1em plus 0.5em minus
  0.4em\relax IEEE, 2015, pp. 1--8.

\bibitem{garrido2014automatic}
S.~Garrido-Jurado, R.~Mu{\~n}oz-Salinas, F.~J. Madrid-Cuevas, and M.~J.
  Mar{\'\i}n-Jim{\'e}nez, ``Automatic generation and detection of highly
  reliable fiducial markers under occlusion,'' \emph{Pattern Recognition},
  vol.~47, no.~6, pp. 2280--2292, 2014.

\bibitem{martcheva2015introduction}
M.~Martcheva, \emph{An introduction to mathematical epidemiology}.\hskip 1em
  plus 0.5em minus 0.4em\relax Springer, 2015, vol.~61.

\bibitem{ames2019control}
A.~D. Ames, S.~Coogan, M.~Egerstedt, G.~Notomista, K.~Sreenath, and P.~Tabuada,
  ``Control barrier functions: Theory and applications,'' in \emph{2019 18th
  European control conference (ECC)}.\hskip 1em plus 0.5em minus 0.4em\relax
  IEEE, 2019, pp. 3420--3431.

\bibitem{wilson2020robotarium}
S.~Wilson, P.~Glotfelter, L.~Wang, S.~Mayya, G.~Notomista, M.~Mote, and
  M.~Egerstedt, ``The robotarium: Globally impactful opportunities, challenges,
  and lessons learned in remote-access, distributed control of multirobot
  systems,'' \emph{IEEE Control Systems Magazine}, vol.~40, no.~1, pp. 26--44,
  2020.

\end{thebibliography}

\end{document}